%% file: main.tex
\documentclass{article}

\usepackage{microtype}
\usepackage{graphicx}
\usepackage{subcaption}
\usepackage{booktabs} 

\usepackage{hyperref}

\usepackage[preprint]{icml2026}

\usepackage{amsmath}
\usepackage{amssymb}
\usepackage{mathtools}
\usepackage{amsthm}

\usepackage[capitalize,noabbrev]{cleveref}

\usepackage{enumitem}
\usepackage{multirow}
\usepackage{xcolor}
\definecolor{darkgreen}{RGB}{0,100,0}
\definecolor{mediumgreen}{RGB}{0,128,0}
\definecolor{lightgreen}{RGB}{144,238,144}
\definecolor{limegreen}{RGB}{50,205,50}
\definecolor{forestgreen}{RGB}{34,139,34}

\theoremstyle{plain}

\theoremstyle{definition}

\theoremstyle{remark}

\newcommand{\ours}{MACF}

\usepackage[textsize=tiny]{todonotes}

\icmltitlerunning{Submission and Formatting Instructions for ICML 2026}

\begin{document}

\twocolumn[
  \icmltitle{Scaling Video Understanding via Compact Latent Multi-Agent Collaboration}
  \icmlsetsymbol{equal}{*}
  \icmlsetsymbol{intern}{$\dagger$}

  \begin{icmlauthorlist}
    \icmlauthor{Kerui Chen}{equal,intern,comp,yyy}
    \icmlauthor{Jinglu Wang}{equal,comp}
    \icmlauthor{Jianrong Zhang}{yyy}
    \icmlauthor{Ming Li}{sch}
    \icmlauthor{Yan Lu}{comp}
    \icmlauthor{Hehe Fan}{yyy}
  \end{icmlauthorlist}

  \icmlaffiliation{comp}{Microsoft Research Asia}
  \icmlaffiliation{yyy}{Zhejiang University}
  \icmlaffiliation{sch}{Guanming Lab}

  \icmlcorrespondingauthor{Hehe Fan}{hehefan@zju.edu.cn}
  \icmlcorrespondingauthor{Yan Lu}{yanlu@microsoft.com}
  \icmlkeywords{Machine Learning, ICML}

  \vskip 0.3in
]

\printAffiliationsAndNotice{\icmlEqualContribution\textsuperscript{$\dagger$}Work done when the author was an intern at Microsoft Research Asia. }

\input{sec/abstract}
\input{sec/intro1}
\input{sec/related}

\input{sec/method}
\input{sec/experiment1}
\input{sec/conclusion}
\input{sec/impact}
\newpage
\nocite{langley00}

\bibliography{example_paper}
\bibliographystyle{icml2026}

\end{document}

%% file: sec/abstract.tex
\begin{abstract}
Multi-modal large language models (MLLMs) advance vision–language understanding but face inherent limitations in long-video tasks due to bounded perception context budgets.
Existing agentic methods mitigate this via rule-based preprocessing, yet often suffer from information loss, high cost, and reliance on textual intermediates.
We propose \ours, an end-to-end Multi-Agent Collaboration Framework that decouples per-agent perception budgets from global video complexity, enabling scalable video understanding while preserving visual fidelity. \ours~partitions videos into segments for locally budgeted agents and enables holistic reasoning via an agent-native latent communication protocol. Each agent encodes partial observations into compact, task-sufficient tokens in a shared embedding space, allowing efficient and information-preserving collaboration by a central coordinator.
We introduce a curriculum training strategy that progressively enforces semantic alignment, evidence summarization, and cross-agent coordination. Extensive experiments on diverse video understanding benchmarks show that \ours~consistently outperforms state-of-the-art MLLMs and multi-agent systems under identical budget constraints, demonstrating the effectiveness of our latent collaboration for scalable video understanding.
\end{abstract}

%% file: sec/intro1.tex
\section{Introduction}
\begin{figure}[!t]
\centering
\includegraphics[width=1.0\linewidth]{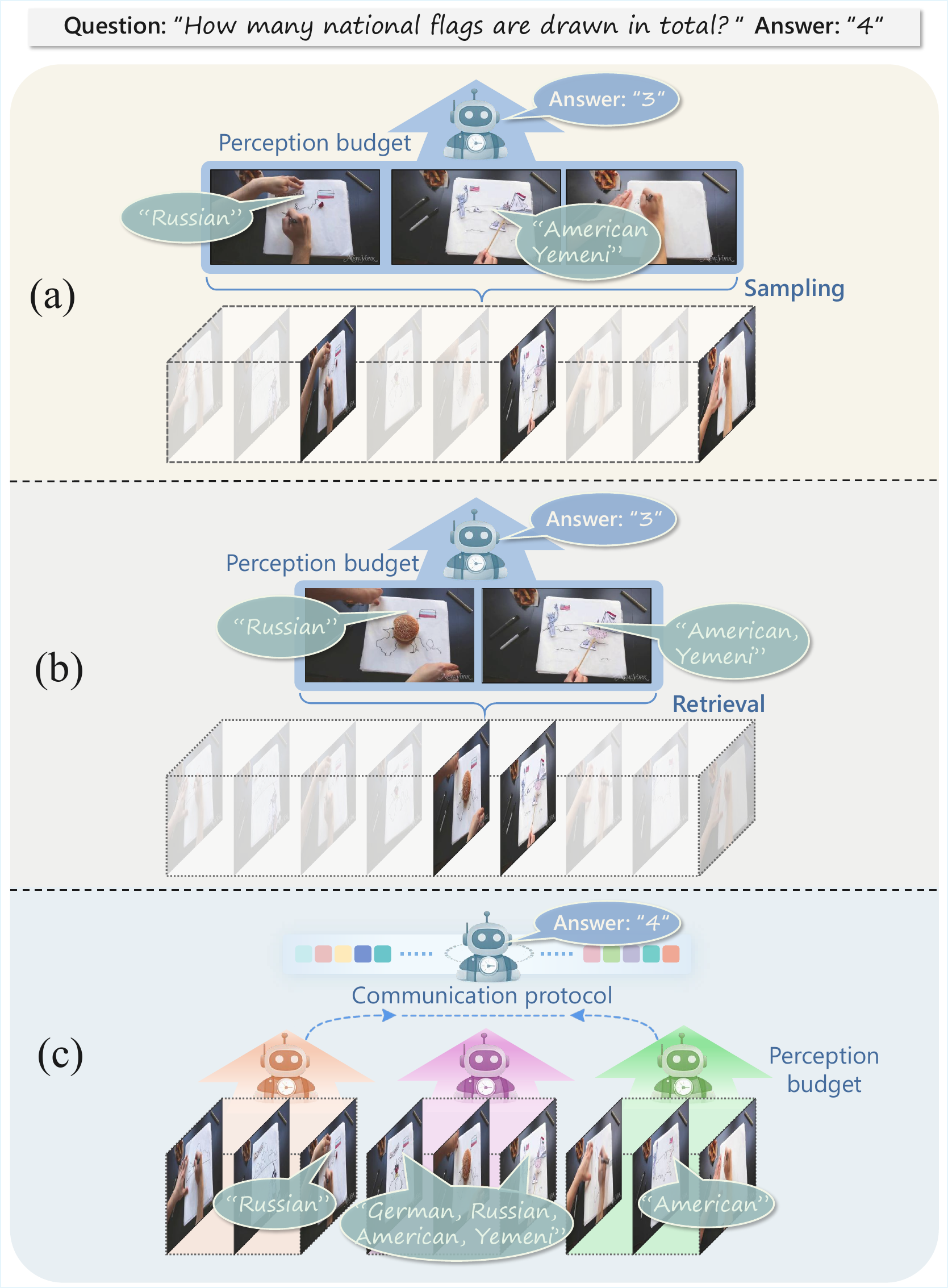}
\vspace{-6mm}
\caption{\textbf{Scaling video understanding with budgeted agents.} (a) Perceptual sampling meets the budget but discards temporal and spatial details (e.g., missing the ``German flag''). 
(b) Retrieval selects key frames based on captions, incurring high cost and potential loss of visual fidelity. 
(c) \ours~partitions the video across local agents and communicates compact latent tokens in a shared space for coordination. (Text hints in green bubbles are for illustration only.)}
\vspace{-6mm}
\label{fig:intro}
\end{figure}

Multi-modal large language models (MLLMs)~\cite{alayrac2022flamingo, li2023blip, gpt-4o, liu2023visual} have demonstrated remarkable capability in vision-language understanding~\cite{yin2024survey, caffagni2024revolution, zhang2024mm}. However, a single MLLM inherently remains constrained by limited input context budgets, hindering scalability to long-temporal or high-resolution perceptual inputs. This limitation is especially evident in video understanding, where raw visual streams are highly redundant, less information-dense than language, and weakly aligned with tasks, causing perceptual inputs to dominate the budget while contributing inefficiently to reasoning.

To address this limitation, agentic methods \cite{yao2025survey, zhou2025reagent, yu2024frame, videorag, long2025seeing} augment MLLMs with auxiliary functions such as preprocessing, retrieval, and memory. A common preprocessing approach is perceptual sampling \cite{han2024self, liang2024end}, which temporally and spatially downsamples videos to meet the input budget. Such approaches inevitably sacrifice temporal continuity and spatial fidelity, leading to substantial information loss, as illustrated in \cref{fig:intro} (a). More advanced systems \cite{fan2024videoagent, wang2024videoagent} rely on retrieval to select a subset of ``informative'' frames, typically guided by caption-based matching, as illustrated in \cref{fig:intro} (b). However, they incur high cost, depend heavily on intermediate text quality, and risking loss of visual details. 
Overall, these pipelines remain rule-driven and text-centric, limiting end-to-end alignment with native visual representations.

A natural solution is to distribute the perceptual workload across multiple agents, each under a local budget constraint, and to aggregate their partial knowledge through collaboration, as shown in \cref{fig:intro} (c). Related paradigms, such as coarse-to-fine \cite{wang2024videoagent} or MapReduce \cite{MapReduce} schemes, follow this intuition but still rely on expensive retrieval pipelines and textual communication across agents. Consequently, this distributed setup shift challenges to the inter-agent communication protocol. It must address:
(1) \textit{Information fragmentation}, requiring aggregation across agents due to partial observations;
(2) \textit{Communication bandwidth constraints}, requiring exchanged messages to be compact; and
(3) \textit{Task-relevant information preservation}, ensuring minimal semantic loss during communication.
Although text-based communication is commonly adopted \cite{zuo2025videolucy, kugo2025videomultiagents, liu2025longvideoagent}, it struggles to convey fine-grained visual cues (e.g., motion, spatial relations, and textures) critical for video understanding. These limitations motivate the need for agent-native communication mechanisms that operate directly in the visual-semantic representation space~\cite{sukhbaatar2016learning, foerster2016learning}.

In this work, we propose an end-to-end Multi-Agent Collaboration Framework (\ours) that scales video understanding by decoupling per-agent perception budgets from global video complexity. 
\ours~partitions the input video into multiple segments, each processed by a local agent within its perception budget. Our key innovation lies in a latent communication protocol that enables efficient and expressive cross-agent information exchange. Specifically, each local agent first maps its observations into compact, task-relevant communication tokens embedded in a shared continuous space. These agent-native tokens form a unified communication channel that preserves visual semantics often lost in textual descriptions. A central coordinator agent ingests tokens from all local agents under a communication bandwidth constraint and performs global understanding. As all messages reside in the same embedding space, cross-agent communication is both efficient and semantically aligned.
To enable effective optimization, we introduce a three-stage curriculum training strategy that progressively establishes semantic alignment, query-aware evidence summarization, and cross-agent collaboration within the shared communication space. Extensive experiments on standard video understanding benchmarks demonstrate that \ours~consistently outperforms state-of-the-art MLLMs under identical budget constraints, validating the effectiveness of agent-native latent communication for scalable video understanding.

In summary, our contribution lies in three-fold.
\begin{itemize}[leftmargin=*, nosep]
    \item We formalize a budget-constrained multi-agent video understanding problem and propose \ours, an end-to-end distributed framework that scales video understanding by decomposing global perception into locally budgeted agents with coordinated inference.
    \item We introduce an agent-native latent communication mechanism that transforms partial visual observations into compact, task-sufficient representations in a shared embedding space, enabling bandwidth-efficient and information-preserving aggregation across agents.
    \item We design a curriculum learning strategy that progressively enforces semantic alignment and coordination in the latent communication space, stabilizing optimization and enabling reliable global inference from fragmented, budget-limited evidence.
\end{itemize}

%% file: sec/related.tex
\section{Related Work}
\noindent\textbf{Multimodal large language models for Understanding.}
Recent multimodal large language models (MLLMs) align pretrained LLMs with visual encoders to support instruction-following, grounding, and open-ended understanding over images~\cite{alayrac2022flamingo, li2023blip, liu2023visual, achiam2023gpt}.
Building on this paradigm, video-oriented MLLMs extend the visual front-end with temporal modeling and frame sampling to answer compositional queries that require motion cues and long-range temporal dependencies~\cite{li2025videochat, zhang2023video, chen2023videollm}.
However, practical deployments face strict input-bandwidth or context limits, where aggressively downsampling frames or shortening clips can discard fine-grained evidence and break temporal reasoning in long videos~\cite{wu2024longvideobench, wang2025lvbench}.
Our work targets this bottleneck by distributing perception across multiple segment-level VLM agents and aggregating their evidence by a coordinator.

\noindent\textbf{Multi-agent communication.}
Most existing LLM-based agent frameworks communicate via free-form natural language~\cite{wu2024autogen, li2023camel, qian2024chatdev, hong2023metagpt}, which has proven effective for a wide spectrum of language tasks such as planning, coding, multi-step reasoning, and tool orchestration.
These agentic paradigms have also been extended to multimodal tasks~\cite{fan2024videoagent, wu2023visual, shen2023hugginggpt, yang2023mm, MapReduce, dvd} by pairing an LLM coordinator or planner with specialized perception VLMs, enabling visual reasoning and content understanding in images and videos.
However, when tasks require fine-grained visual evidence and temporal relations, discrete language communication becomes a bottleneck because not all visual evidence can be expressed precisely~\cite{su2025thinking}, and summarization often drop critical spatial, grounding and temporal cues.
Beyond text messaging, emerging efforts in language domain~\cite{LatentMAS,fu2025cache} explore latent or state-based coordination (e.g., transferring internal states such as KV-cache) to increase communication bandwidth and efficiency.
In this paper, we study distributed multimodal evidence communication, where agents write visual evidence into a shared communication space and a coordinator integrates these distributed information for global answering.

\noindent\textbf{Context compression.}
Beyond parameter compression~\cite{frantar2022gptq, mozaffari2024slim}, recent studies explore compressing long contexts into compact latent tokens that can be directly conditioned on by an LLM, alleviating context-length and inference-memory bottlenecks~\cite{ge2023context, he2025clara}.
Such memory-slot based compression provides a model-native interface for retaining evidence under a fixed token budget, avoiding information loss caused by the natural language.
Different from prior context compression which targets single model language long-context usage, we focus on \emph{multi-agent} settings and compress each agent's \emph{multimodal observations} into fixed-length communication tokens for communicating and downstream evidence aggregation.
This design explicitly treats the latent tokens as a communication channel, enabling the coordinator to compose distributed visual evidence across temporal segments.

\begin{figure*}
    \centering
    \includegraphics[width=0.96\linewidth]{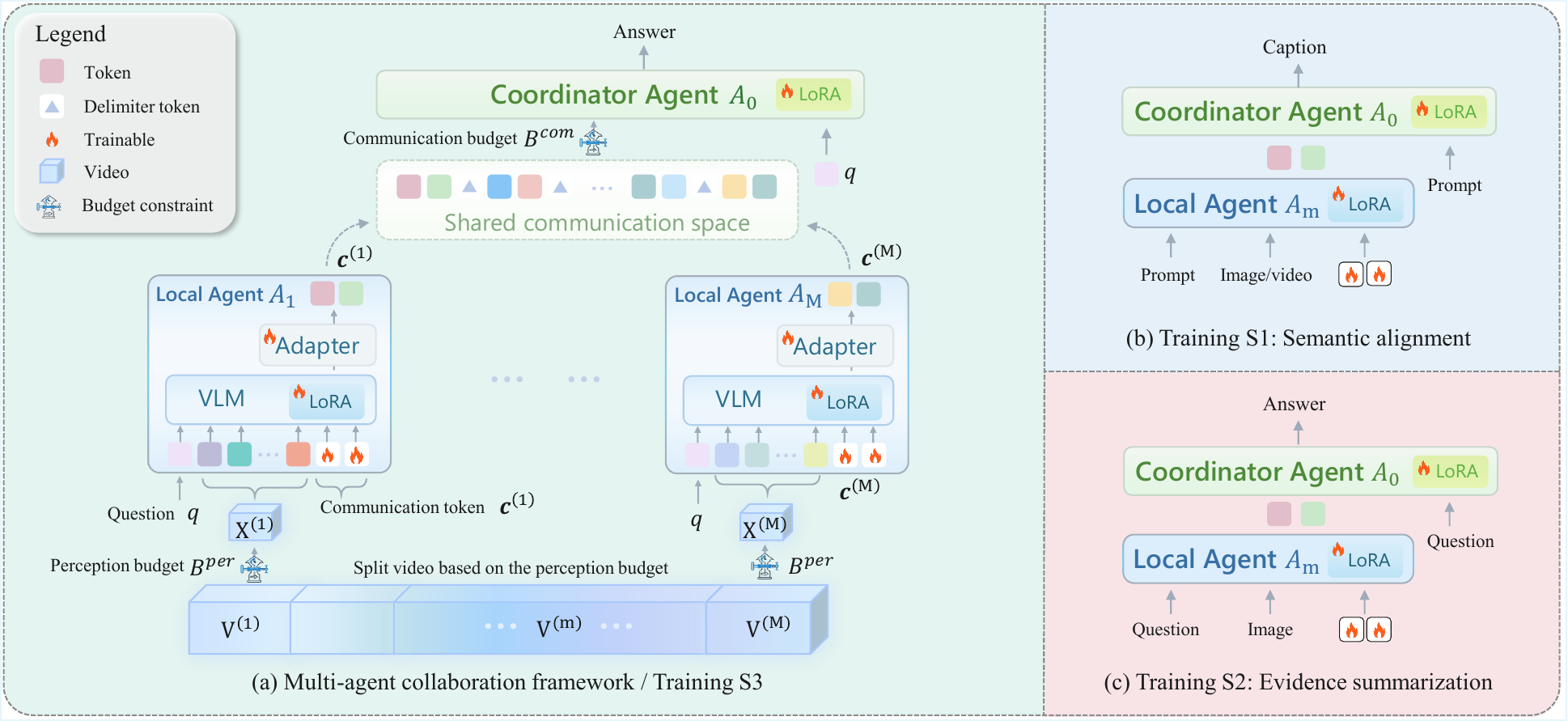}
    \caption{\textbf{Overview of MACF.} 
    (a) \ours~distributes the video perception workload temporally across multiple local agents, each operating under a per-agent perception budget. Each agent compresses its observation into communication tokens extracted from the last-layer hidden states and projects them into a shared latent communication space via an adaptor module. A coordinator agent aggregates tokens from all agents to produce the final prediction. (b) and (c) illustrate the curriculum training, which includes semantic alignment and evidence summarization, followed by end-to-end coordination training shown in (a).
    }
    \label{fig:pipeline}
    \vspace{-3mm}
\end{figure*}

%% file: sec/method.tex
\section{Method}

\subsection{Problem Formulation}
To formalize scalable video understanding in real-world scenarios, we consider the problem formulation in which each agent operates under explicit budget constraints on both perception and communication.

Let $V=\{I_i\}_{i=1}^T$ denote an input video consisting of $T$ frames, where each frame $I_i$ with original resolution $H \times W$.
Given a textual query $q$, the objective is to predict an answer $y$, which may be a categorical label, a free-form textual response, or a structured output, depending on the downstream task.

\noindent\textbf{Perception budget.}
In practical deployments, each agent can process only a bounded amount of visual input per forward pass.
We formalize this bound as a perception budget $B^{per}$, defined as the maximum number of pixels that can be ingested in a single inference.
This pixel-based budget offers a simple and unified proxy for visual input capacity across different systems, models, or codec algorithms.
A single agent is responsible for processing the video $V$ under the perception budget:
\begin{equation}
\label{eq:perception_budget}
    \mathrm{pix}(X) \leq B^{per}, \quad X=\mathcal{P}(V),
\end{equation}
where $\mathrm{pix}(\cdot)$ counts the number of pixels in the processed input, $\mathcal{P}(\cdot)$ denotes a preprocessing function that may include temporal and spatial subsampling, including a trade-off between temporal coverage and spatial fidelity.

To overcome this perception bottleneck, we can distribute the perceptual workload across multiple agents.
Given a set of $M$ agents $\mathcal{A} = \{A_1, ..., A_M\}$, each $A_m$ perceives a video segment $V^{(m)}$ via a local transformation $X^{(m)} = \mathcal{P}_m (V^{(m)})$, subject to a per-agent pixel budget, $ \mathrm{pix}(X^{(m)}) \leq B^{per}$.
This decouples the global video length $T$ from the per-agent input budget $B^{per}$, allowing the system to scale to substantially longer videos while preserving fine-grained visual information within each local segment.

\noindent\textbf{Communication budget.}
Since no single agent observes the full video, agents must communicate local information for global coordination. This process is constrained by the finite context capacity of the coordinator $A_0$, which we model as a communication budget $B^{com}$ representing the maximum information it can ingest per forward pass.
Let $\mathbf{c}^{(m)}$ denote the communication message produced by agent $A_m$. The coordinator $A_0$ consumes a subset of the messages $\mathcal{C} \subseteq \{\mathbf{c}^{(1)},...,\mathbf{c}^{(M)}\}$ from all agents.
The constraint is:

\begin{equation}
\sum_{\mathbf{c}^{(m)} \in \mathcal{C}} \big|\mathbf{c}^{(m)}\big| + |q|
\;\le\;
B^{com}.
\end{equation}

\subsection{Multi-Agent Collaboration Framework (\ours)}
We introduce the Multi-Agent Collaboration Framework (\ours) for scalable video understanding with compact latent collaboration.
As illustrated in \cref{fig:pipeline} (a), \ours~consists of three components: 1) distributed local agents with budgeted perception; 2) compact latent communication tokens for information exchange; 3) a coordinator agent fusing distributed evidence for global understanding.
The entire framework is fully differentiable and trained end-to-end.

\paragraph{Local agents with budgeted perception}
Without loss of generality, we assume all local agents share the same architecture and operate under an identical perception budget.
Given $M$ available agents $\{A\}_{m=1}^M$, the input video is first partitioned into disjoint temporal segments $V = \{V^{(1)}, V^{(2)}, \dots, V^{(M)}\}$, where each segment $V^{(m)}$ is assigned to agent $A_m$, for $m=1,\dots,M$.
Focusing on multi-agent collaboration rather than sampling or compression strategies, we adopt a simple and consistent preprocessing scheme: each agent uniformly samples $F$ frames from its segment and resizes them to a shared resolution $(h, w)$. The resulting input satisfies the per-agent budget:
\vspace{-1mm}
\begin{equation}
    F \times h \times w \leq B^{per}.
\end{equation}
\vspace{-1mm}

Scaling the number of agents effectively redistributes the perception budget across time, allowing finer-grained spatiotemporal representation within each local segment under a fixed per-agent constraint.

\noindent\textbf{Latent collaboration with communication tokens.}
Given the uniform video partitioning, each agent’s local observation can contribute to the global task. We adopt a star-topology in which all local agents communicate directly with a central coordinator. Unlike prior multi-agent systems that rely on natural language for inter-agent communication~\cite{MapReduce}, we avoid text-based messaging, which is often redundant and lossy for video understanding.

We propose to communicate using latent communication tokens.
A local agent summarizes its visual evidence into compact continuous vectors encoding query-relevant evidence softly.
These \textit{communication tokens} are information-dense and directly consumable by the coordinator, enabling efficient and faithful collaboration in a shared representation space.
Formally, each agent $A_m$ encodes its local observation $X^{(m)}$ and query $q$ into a fixed-capacity communication token sequence:
\begin{equation}
    \mathbf{c}^{(m)} = A_m(X^{(m)}, q) \in \mathbb{R}^{K \times d},
\end{equation}
where $K$ is the number of communication tokens per agent and $d$ is the embedding dimension.

In the star-topology, the coordinator $A_0$ aggregates tokens from all agents to produce the final prediction:
\begin{equation}
\label{eq:predict}
    \hat{y} = A_0 (q,  \mathbf{c}^{(1)},...,  \mathbf{c}^{(M)}).
\end{equation}
Communication is constrained by the coordinator’s finite context capacity, modeled as a communication budget measured in tokens:
\begin{equation}
    M \times K + \mathrm{token}(q) \;\le\;
B^{com},
\end{equation}
where $\mathrm{token}(\cdot)$ denotes number of tokens.

\subsection{Optimization of \ours}
Since all local agents ${A}_{m=1}^M$ share the same input and output formats and model architectures, we tie their parameters and learn a single set of shared weights during training. This design reduces model complexity and enforces consistent behavior across agents processing different video segments. Notably, our framework can be extended to heterogeneous models, as each agent is equipped with an adaptor module that maps its communication tokens into a shared latent space consumable by the coordinator $A_0$.

Under the perception and communication constraints, the objective is to jointly learn the shared local-agent model $A_m$ and the coordinator $A_0$ by minimizing the expected task loss:
\begin{equation}
\min_{\theta_A, \theta_0}
\mathbb{E}_{(V,q,s)\sim\mathcal{D}}
\big[\ell(\hat{y}, y)\big],
\end{equation}
where $\theta_A$ denotes the parameters shared by all local agents and $\theta_0$ the parameters of the cooridnator $A_0$, $\mathcal{D}$ the data distribution of triplets $(V, q, y)$, $\ell(\cdot)$ measuring the discrepancy and often instantiated as autoregressive cross-entropy (CE).

Although \ours~is fully differentiable, direct end-to-end training is challenging due to a two-stage bottleneck that induces an effective min-cut in the information flow and the asymmetric roles of local agents and the coordinator.
Local agents are responsible for mapping high-dimensional video inputs into concise communication tokens in a shared latent space, while the coordinator must aggregate these tokens and perform global coordination across agents.
Training both components simultaneously from scratch often leads to unstable representations and ineffective collaboration.
To address these challenges, we adopt a curriculum training strategy that progressively equips the model with the required capabilities, with three stages:
: (1) semantic alignment to anchor communication tokens in a shared latent space; (2) evidence summarization to encode query-relevant information into compact tokens; and (3) cross-agent collaboration to enable effective global aggregation.

\paragraph{Semantic alignment.}

Before effective collaboration can emerge, communication tokens produced by local agents must be anchored in a unified semantic space that is consumable by the coordinator.
Without such alignment, the communication tokens can drift across agents and lack semantic consistency.
Accordingly, in the first training stage illustrated in \cref{fig:pipeline} (b), we initialize the \emph{communication space} with caption supervision, which provides dense, low-variance language targets and stabilizes the grounding of visual evidence.
Given the preprocessed visual input $X$, a local agent $A_m$ map it into communication tokens $\mathbf{c}=A_m(X)$, and the coordinator $A_0$ decodes the corresponding caption conditioned on these tokens. We optimize the standard autoregressive cross-entropy (CE) loss with respect to the ground-truth caption $C$:
\begin{equation}
\mathcal{L}_{\text{cap}}
= 
\mathrm{CE}\big(C,\; A_0(\mathbf{c})\big),
\label{eq:cap_loss}
\end{equation}
This first stage establishes a unified communication space, serving as the foundation for subsequent query-aware evidence summarization and cross-agent collaboration.

\paragraph{Evidence summarization.}
After semantic alignment, communication tokens lie in a shared latent space, but may not yet encode query-relevant information.
Local agents must therefore learn to select and compress task-relevant evidence under a fixed communication budget.
To this end, we perform supervised fine-tuning (SFT) on a high-quality Image-QA dataset. This stage encourages local agents to encode question-conditioned evidence into compact communication tokens, as illustrated in \cref{fig:pipeline} (c). Given an image $X$ and question $q$, a local agent $A_m$ produces communication tokens $\mathbf{c}=A_m(X, q)$ and the coordinator $A_0$ conditions on both the communication tokens and the query embedding to generate the target answer.
Training is performed using a standard autoregressive cross-entropy loss on the target sequence:
\begin{equation}
\mathcal{L}_{\text{evi}} = \mathrm{CE}\big(y, A_0([\mathbf{c};\mathrm{Emb}(q) ]\big).
\label{eq:evi_loss}
\end{equation}
This stage trains local agents to produce compact, query-aware evidence summaries consumable by the coordinator.

\paragraph{Cross-agent collaboration.}
The ultimate goal of \ours\ is to reason over distributed observations collected by multiple local agents.
At this stage, the coordinator must not only interpret individual communication tokens, but also fuse evidence across agents and perform global coordination, as illustrated in \cref{fig:pipeline} (a).
We train this capability using Video-QA datasets. Following the prediction formulation in \cref{eq:predict}, the final loss takes the form:
\begin{equation}
\mathcal{L}_{\text{col}} = \mathrm{CE} \big(y,A_0([\mathbf{c}^{(1)};...; \mathbf{c}^{(M)}; \mathrm{Emb}(q)])\big).
\label{eq:vid_loss}
\end{equation}

Overall, \ours\ enables scalable video understanding under explict perception and communication budgets by distributing video perception across local agents and exchanging compact latent communication tokens with a coordinator.
A three stage curriculum forms the shared communication space and equips the coordinator to fuse distributed evidence for global reasoning.

%% file: sec/experiment1.tex
\input{tables/main_table}
\section{Experiment}
To validate \ours\ under budget-limited video understanding, we set the perception budget and conduct extensive experiments on video understanding task.
In this section, we first details the setup of experiments.
Then we report main results on four long video benchmarks (\cref{main_results}), followed by analyses on communication, efficiency, and ablations of our training curriculum (\cref{discussion}).

\subsection{Setup}
\noindent\textbf{Implementation.}
We use Qwen3-VL-8B~\cite{qwen3} as the local agents and the coordinator agent in the main results.
During training on video data, each local agent receives $F=16$ frames as input, and each frame is processed at a maximum resolution, $h \times w = 224 \times 224$.
For the final training stage, we partition each video across $M=4$ local agents for distributed understanding under the same per-agent input budge.
Our training proceeds in three stages. 
In Stage~1, we use caption data from the 0-30s subset of LLaVA-Video-178K~\cite{llava-video-178k}.
In Stage~2, we use image data from Video-R1~\cite{video-r1}.
In Stage~3, we train on video data from Video-R1 together with a subset of Molmo2~\cite{clark2026molmo2}.
At inference time, we increase the number of local agents to 6.
All training is conducted on 4$\times$NVIDIA A100 (80GB) GPUs. More details are provided in the appendix.

\noindent\textbf{Baselines.}
We mainly compare our method with thirteen open-source and proprietary state-of-the-art MLLMs: GPT-4o~\cite{gpt-4o}, LLaVA-Next-Video-34B\cite{zhang2024llavanextvideo}, ShareGPT4Video-8B~\cite{chen2024sharegpt4video}, Kangaroo-8B~\cite{liu2024kangaroo}, VideoLLaMA2.1-7B~\cite{videollama2}, VideoLLaMA3-7B~\cite{videollama3}, Qwen2.5-VL-72B~\cite{qwen2.5}, Qwen2.5-VL-7B, LLaVA-OneVision1.5-8B~\cite{an2025llava}, Keye1.5-VL-8B~\cite{keye1.5}, InternVL2.5-VL-8B~\cite{internvl2.5}, Qwen-VL-30B and Qwen3-VL-8B~\cite{qwen3}.
Moreover, to verify the effectiveness of our communication protocol, we implement two baseline and conduct experiments following the setups of MapReduce~\cite{MapReduce} and LatentMAS~\cite{LatentMAS}.
For a fair comparison, we replace the closed-source model used in the original MapReduce implementation with Qwen3-VL-8B. 
We also adapt and modify the official LatentMAS code to support multimodal understanding.
The implementation details of these three baselines are provided in the Appendix.

\noindent\textbf{Benchmarks.}
We evaluate our method on four general-purpose long video understanding benchmarks: Video-MME~\cite{video-mme}, LongVideoBench~\cite{wu2024longvideobench}, LVBench~\cite{wang2025lvbench} and MLVU-Test~\cite{zhou2025mlvu}.
For MLVU-Test, we evaluate on its multiple-choice question set for stability and consistency.
For all evaluations, we follow the decoding configuration used in the official demo code.

\subsection{Main Results}
\label{main_results}
As shown in \cref{main_table}, we compare against baselines on four standard benchmarks, demonstrating the effectiveness of \ours\ for general video understanding. We highlight two main findings below.

\noindent\textbf{Superior performance.}
Under the input budget constraint, many existing methods exhibit noticeable performance degradation. 
In contrast, MACF consistently surpasses previous models across all the benchmarks.
Compared with the Qwen3-VL-8B, MACF achieves consistent gains of \textbf{+4.5\%}, \textbf{+6.1\%}, \textbf{+7.0\%}, and \textbf{+7.7\%} on Video-MME, LongVideoBench, LVBench, and MLVU-Test, respectively.
Notably, on MLVU-Test, MACF surpasses all compared open-source models, even when those baselines are evaluated without input budget constraints.
Overall, these results validate the effectiveness of MACF’s design for budget-limited long-form video understanding.

\noindent\textbf{Generalization to other MLLMs backbone.}
To demonstrate the generalization of our proposed MACF, we conduct experiments with Qwen2.5-VL-7B and LLaVA-OneVision1.5-8B as further validations.
As shown in ~\cref{main_table}, MACF consistently improves performance across all benchmarks when applied to different foundational backbones (e.g., Qwen3, LLaVA, and Qwen2.5). Specifically, our method yields an average absolute improvement of \textbf{+6.3\%} for Qwen3-VL-8B and achieves an overall average gain of approximately \textbf{+5.6\%}.

\input{fig/vis_1}
\noindent\textbf{Effective communication representation.}
We further analyze the effectiveness of our communication protocol by comparing it with text-only communication and a LatentMAS KV-cache sharing baseline under same task setting.
Across Video-MME, LongVideoBench, LVBench, and MLVU-Test, MACF outperforms text-based communication by \textbf{20.3\%}, \textbf{18.6\%}, \textbf{9.7\%}, and \textbf{15.9\%}, respectively.
Compared with directly sharing KV-Caches as in LatentMAS, MACF still achieves consistent improvements of \textbf{9.3\%}, \textbf{14.1\%}, \textbf{10.5\%}, and \textbf{10.3\%} on the same benchmarks.
These results indicate that explicitly designed communication tokens provide a more effective and structured medium for cross-agent information exchange than either textual summarization or raw latent sharing.
It enable the coordinator to aggregate visual evidence more reliably under strict input budget constraints.

As shown in \cref{fig:vis}, we further conduct a qualitative experiment to compare text-based communication with our latent communication framework.
In this example, the question asks for the rope color of the white dog, but the video clips in different agents contain multiple dogs.
Under text-only communication, each agent must discretize its observation into a short caption-like statement.
This compression not only drops visual details, but also easily introduces a reference binding error: an agent (agent6) may mistakenly associate the queried “white dog” with the yellow-white dog in its own segment and report the rope color as red.
Once such error statement enters the shared text context, the coordinator may produces the incorrect answer.
In contrast, MACF’s communication tokens preserve fine-grained visual evidence in a structured latent form, enabling the coordinator to align information across agents more reliably and recover the correct answer.

\input{sec/discussion}

\input{tables/ablation_compression_ratio}
\input{fig/curve1}

\noindent\textbf{Communication cost.}
As shown in \cref{tab:infer}, we report the communication throughout and response latency for different communication methods. 
Our approach achieves the lowest throughout and the shortest response latency.
In contrast, LatentMAS \cite{zou2025latent} incurs the highest communication cost due to the necessity of transmitting high dimensional KV-Caches. 
Meanwhile, MapReduce \cite{MapReduce} wit text-based communication suffers from the highest latency, primarily stemming from the autoregressive decoding process required for each agent interaction.

\noindent\textbf{Effect of training strategy.}
To investigate the effect of our proposed training strategy, we conduct experiments on Video-MME and MLVU.
As shown in the \cref{tab:ablation_stages}, the full training strategy achieves the best performance.
\input{tables/inference_time}
\input{fig/curve2}
Removing any stage leads to substantial performance drops, with average decreases of \textbf{13.9\%}, \textbf{3.0\%}, and \textbf{12.5\%}, respectively.
These ablation results provide three key findings.
First, stage~1 demonstrates that aligning the communication tokens with the coordinator’s representation space forms a well-initialized interface that is important for complex QA.
Second, stage~2 shows that strengthening high-quality image QA understanding substantially benefits video understanding.
Third, stage~3 highlights the importance of learning cross-agent collaboration for effective distributed communication.
\input{tables/ablation_stage}

\input{tables/model_agnostic}
\noindent\textbf{Model agnostic.}
To demonstrate the model-agnostic nature of MACF, we conducted experiments with a fixed coordinator backbone (Qwen3-VL-8B) and different local agent backbones.
Specifically, we evaluated two distinct configurations: one where all local agents utilize Qwen2.5-VL-7B, and another employing a heterogeneous mixture of Qwen2.5-VL-7B and LLaVA-OneVision1.5-8B.
As shown in~\cref{tab:model_agnostic}, the results confirm that MACF generalizes beyond a single model family. This is enabled by the adapter module (2-layer MLP), which projects any backbone's hidden states into a shared latent space with fixed dimensionality $c^{(m)}\in \mathbb{R}^{K\times d}$, absorbing architectural differences across backbones.

%% file: tables/main_table.tex
\begin{table*}[t!]
\centering
\resizebox{0.9\linewidth}{!}{
\setlength{\tabcolsep}{10pt}
\begin{tabular}{c|ccccc}
\toprule
{Model} & {Video-MME} & {LongVideoBench} & {LVBench} & {MLVU-Test} & {Perc. budget} \\
\midrule
\multicolumn{6}{c}{\textit{MLLMs without perception budget constraint}} \\
\midrule
GPT-4o & 71.9 & 66.7 & 30.8 & 54.9 & - \\
LLaVA-Next-Video-34B & 52.0 & 50.5 & 32.2 & - & -\\
ShareGPT4Video-8B & 39.9 & 41.8 & - & 33.8 & - \\
Kangaroo-8B & 56.0 & 54.8 & 39.4 & 46.5 & - \\
VideoLLaMA2.1-7B & 54.9 & - & 36.2 & 45.6 & - \\
VideoLLaMA3-7B & 66.2 & 59.8 & 45.3 & 47.7 & - \\
\midrule
\multicolumn{6}{c}{\textit{MLLMs with perception budget constraint}} \\
\midrule
Qwen2.5-VL-72B & {59.5} & 51.0 & 35.7 & {44.2} & 16*224*224\\
Qwen3-VL-30B & 58.3 & 55.9 & {38.0} & 44.0 & 16*224*224\\
Qwen2.5-VL-7B & 53.7 & 48.1 & 32.2 & 38.7 & 16*224*224\\
LLaVA-OneVision1.5-8B & 56.1 & 54.4 & 36.4 & 41.8 & 16*224*224\\ 
Keye1.5-VL-8B & 55.6 & {56.4} & 37.6 & 41.6 & 16*224*224\\
InternVL2.5-VL-8B & 47.7 & 43.4 & 32.3 & 39.1 & 16*224*224\\
Qwen3-VL-8B & 55.9 & 50.7 & 33.2 & 41.5 & 16*224*224 \\
\midrule
\multicolumn{6}{c}{\textit{Multi-agent systems with perception budget constraint}} \\
\midrule
MapReduce$^{*}$ & 46.7 & 38.2 & 30.5 & 33.3 & 16*224*224 \\
LatentMAS$^{*}$ & 55.7 & 48.5 & 33.2 & 42.3 & 16*224*224 \\
Ours (Qwen2.5-VL-7B) & {58.2} & {52.5} & {37.7} & {47.6} & 16*224*224\\
Ours (Qwen3-VL-8B) & \textbf{60.4} & \underline{56.8} & \underline{40.2} & \underline{49.2} & 16*224*224\\
Ours (LLaVA-OV1.5-8B) & \underline{59.9} & \textbf{57.6} & \textbf{41.1} & \textbf{49.4} & 16*224*224\\
\bottomrule
\end{tabular}
}
\caption{Quantitative comparisons of the state-of-the-art MLLMs on the diverse general video understanding benchmark. The best and second best results are \textbf{bold} and \underline{underlined}. All the numbers are presented in \%. The perception budget is specified as the sampled frame number times processing resolution ($F \times h \times w$). * denotes that MapReduce and LatentMAS use Qwen3-VL-8B as the VLM engine.}
\vspace{-6mm}
\label{main_table}
\end{table*}

%% file: fig/vis_1.tex
\begin{figure}[t]
    \centering
    \includegraphics[width=0.95\linewidth]{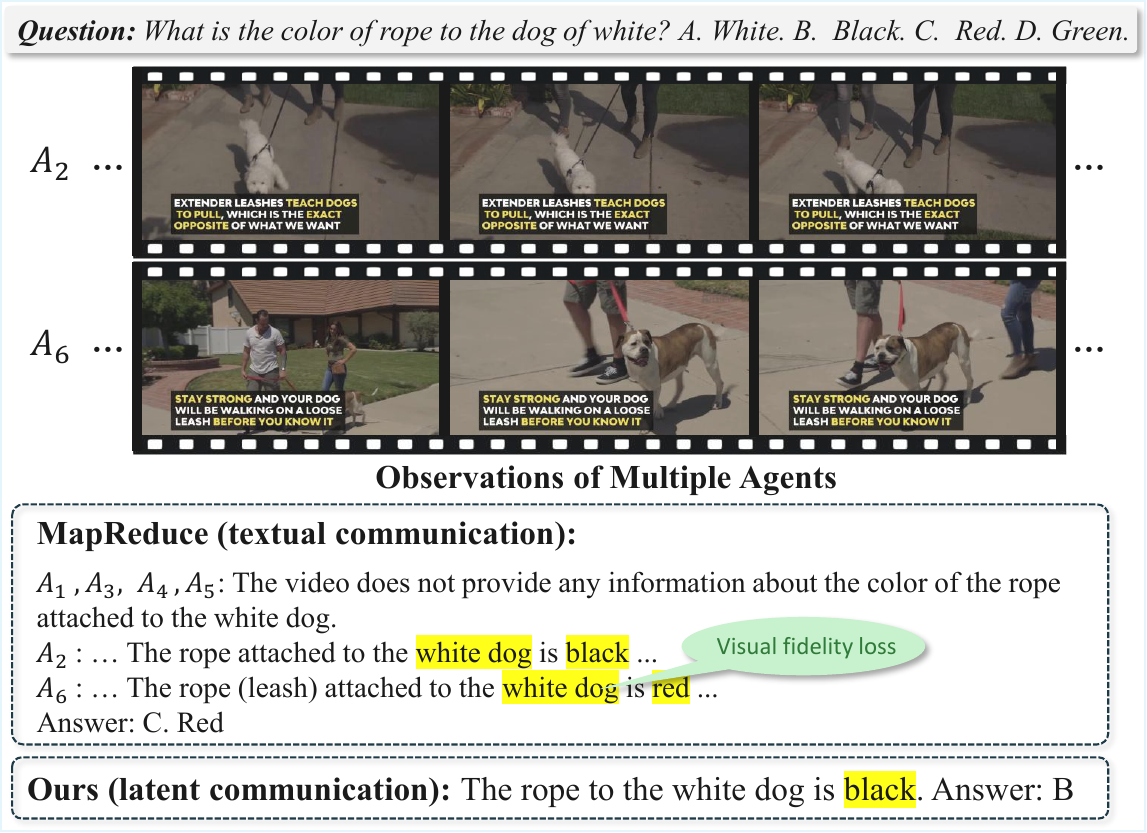}
    \caption{\textbf{Qualitative comparison of communication representations.} Text-based communication (MapReduce) and our latent communication. Text-based communication can fail due to lossy visual descriptions: the ``brwon-and-white dog'' is reduced to ``white dog'', causing Agent 6 to infer an incorrect rope color (red), which the coordinator selects, leading to a wrong answer.  In contrast, our latent communication preserves fine-grained visual evidence and yields the correct result.}
    \label{fig:vis}
    \vspace{-5mm}
\end{figure}

%% file: sec/discussion.tex
\subsection{Discussion}
\label{discussion}

Our framework is fundamentally governed by two information bottlenecks: a \emph{perception bottleneck} at the input level and a \emph{communication bottleneck} at the coordination level. From an information-theoretic perspective, these bottlenecks jointly determine how much task-relevant information can be preserved and transmitted under limited perceptual and communication budgets, and therefore control the system’s ability to scale to long and complex videos.

\noindent\textbf{Perception bottleneck.}
The perception bottleneck characterizes the total amount of visual information that can be processed by the system through all local agents. It is controlled by two complementary factors: temporal and spatial capacity. 
The total perception volume can be measured by the number of processed pixels, $(M \times F) \times (h \times w)$, where $M$ is the number of agents, $F$ the number of frames per agent, and $h \times w$ the input frame resolution. Here, $(M \times F)$ represents temporal capacity, while $(h \times w)$ corresponds to spatial capacity.

\noindent\textbf{(i) Effect of temporal capacity.}
Temporal capacity ($M \times F$) is determined by the number of local agents $M$ when the per-agent temporal input budget (i.e., the number of frames processed by each agent $F$) is fixed. Increasing the number of local agents linearly expands the effective temporal coverage of the video, thereby reducing information loss caused by sparse temporal sampling. We study this effect on MLVU-Test by progressively scaling the number of local agents. As shown in \cref{fig:curve_1}, \ours~exhibits a consistent performance improvement as the number of agents increases.

\noindent\textbf{(ii) Effect of spatial capacity.}
Spatial capacity is determined by the spatial resolution ($h \times  w$) of frames processed by each agent, which directly affects the fidelity of local visual representations. To isolate this factor, we fix both the number of local agents and the number of frames processed per agent, and increase the input bandwidth by allocating higher spatial resolution per frame. As shown in \cref{fig:curve_2}, \ours~exhibits a monotonic performance improvement as resolution increases, indicating reduced spatial distortion in local observations. However, as illustrated by the blue curve with $K=32$ communication tokens, this improvement saturates beyond a certain point once the communication bottleneck, measured by $K \times M$ (see \cref{sec:exp_communication_bottleneck}), becomes limiting.
Notably, increasing the communication budget to $K=48$ (orange curve) raises the performance ceiling, suggesting that higher-fidelity local representations require greater communication capacity for effective aggregation by the coordinator.

\noindent\textbf{Communication bottleneck.}
\label{sec:exp_communication_bottleneck}
The communication bottleneck characterizes the information transmission constraint between local agents $\{A\}_{m=1}^M$ and the coordinator $A_0$. The total communication capacity is bounded by $K \times M + \mathrm{token}(q)$, where $M$ is the number of agents and $K$ the number of communication tokens per local agent (we omit $\mathrm{token}(q)$ as it is data-dependent). For fixed $M$, this bottleneck is governed by $K$ and can be viewed as a \emph{rate constraint} on the information available to the global inference module.
Under the main experimental setting wit fixed $M$, we evaluate communication budgets of $K \in \{16, 32, 48\}$.
As shown in \cref{tab:ratio}, $K=16$ results in substantial performance degradation, indicating excessive compression and loss of task-relevant information. Increasing the budget to $K=32$ yields a significant improvement, while further increasing to $K=48$ provides only marginal gains.
This saturation behavior suggests that once the communication rate exceeds the task-relevant information of the perceived observations, additional bandwidth offers diminishing returns, a hallmark of information bottleneck regimes.

Together, these two bottlenecks define a coupled rate-distortion trade-off between local perception and global aggregation. Insufficient perception bandwidth increases distortion at the input level, while insufficient communication capacity prevents the coordinator from reconstructing a task-sufficient global representation. By jointly controlling these two constraints, \ours~enables scalable video understanding that approaches an efficient operating point under strict perception and communication budgets.

%% file: tables/ablation_compression_ratio.tex
\begin{table}[!t]
  \centering
  \resizebox{0.9\linewidth}{!}{
  \setlength{\tabcolsep}{8pt}
  \renewcommand{\arraystretch}{1.15}
  \begin{tabular}{c|ccc}
    \toprule
    $K$ & MLVU & LVBench & LongVideoBench\\
    \midrule
    16 & 46.0 & 38.7 & 54.1\\
    32 & 49.2 & 40.2 & 56.8\\
    48 & 49.0 & 40.1 & 57.2\\
    \bottomrule
  \end{tabular}
  }
  \caption{Effect of communication bottleneck.}
  \label{tab:ratio}
  \vspace{-6mm}
\end{table}

%% file: fig/curve1.tex
\begin{figure}[!t]
    \centering
    \includegraphics[width=0.9\linewidth]{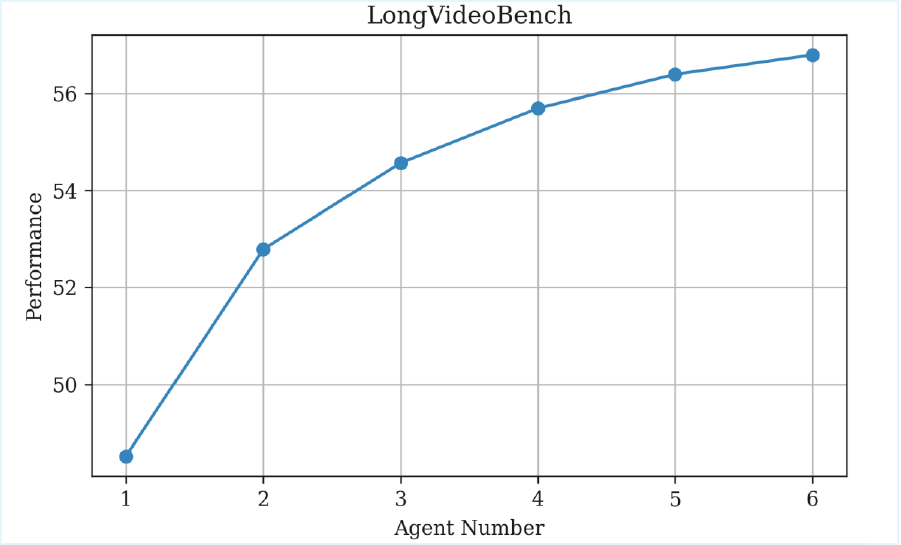}
    \caption{Effect on local agent number $M$. Note that we set $M=4$ in training.}
    \label{fig:curve_1}
    \vspace{-5mm}
\end{figure}

%% file: tables/inference_time.tex
\begin{table}[t]
  \centering
  \resizebox{0.9\linewidth}{!}{
  \setlength{\tabcolsep}{8pt}
  \renewcommand{\arraystretch}{1.15}
  \begin{tabular}{c|cc}
    \toprule
    Method & Inference Time & Throughput\\
    \midrule
    MapReduce & 5.156s & 387 tokens\\
    LatentMAS & 0.649s & 784 KV-Cache\\
    Ours & 0.537s & 192 tokens\\
    \bottomrule
  \end{tabular}
  }
  \caption{Analysis of communication cost.}
  \vspace{-6mm}
  \label{tab:infer}
\end{table}

%% file: fig/curve2.tex
\begin{figure}[!t]
    \centering
    \includegraphics[width=0.9\linewidth]{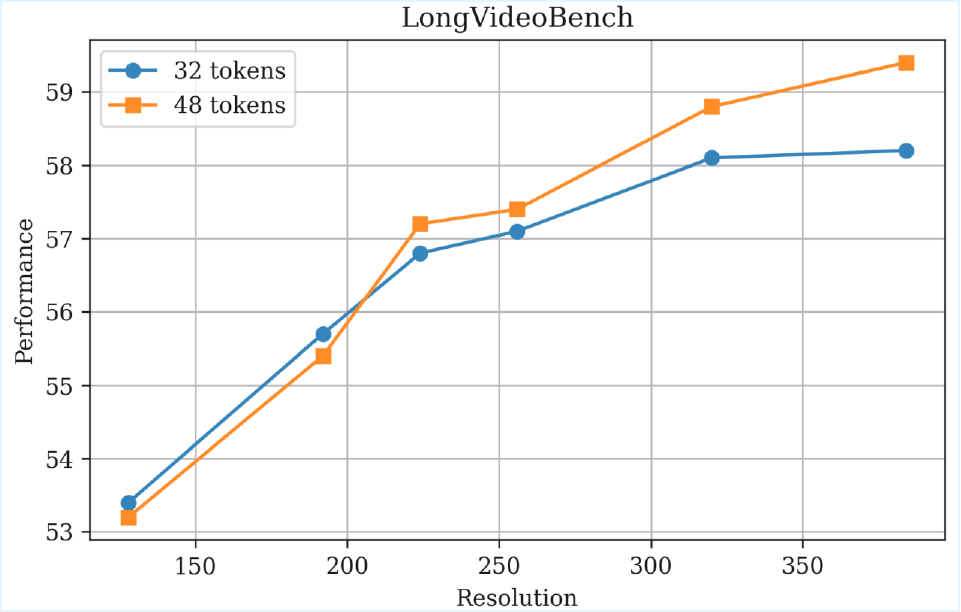}
    \caption{Effect of spatial capacity on LongVideoBench.}
    \label{fig:curve_2}
    \vspace{-4mm}
\end{figure}

%% file: tables/ablation_stage.tex
\begin{table}[!htbp]
  \centering
  \resizebox{0.9\linewidth}{!}{
  \setlength{\tabcolsep}{7pt}
  \renewcommand{\arraystretch}{1.15}
  \begin{tabular}{ccc|cc}
    \toprule
    Stage1 & Stage2 & Stage3 & Video-MME & MLVU \\
    \midrule
    $\times$ & $\checkmark$ & $\checkmark$ & 45.5 & 36.3 \\
    $\checkmark$ & $\times$ & $\checkmark$ & 56.9 & 46.7 \\
    $\checkmark$ & $\checkmark$ & $\times$ & 48.2 & 36.5 \\
    $\checkmark$ & $\checkmark$ & $\checkmark$ & 60.4 & 49.2 \\
    \bottomrule
  \end{tabular}
  }
  \caption{Effect of training stages.}
  \label{tab:ablation_stages}
  \vspace{-8mm}
\end{table}

%% file: tables/model_agnostic.tex
\begin{table*}[t]
  \centering
  \resizebox{0.9\linewidth}{!}{
  \setlength{\tabcolsep}{7pt}
  \renewcommand{\arraystretch}{1.15}
  \begin{tabular}{c|cccc}
    \toprule
    Local Agent & Video-MME & LongVideoBench & LVBench & MLVU-Test \\
    \midrule
    Qwen3-VL-8B & 60.4 & 56.8 & 40.2 & 49.2\\
    Qwen2.5-VL-7B & 56.4 & 51.1 & 36.5 & 47.8\\
    Qwen2.5-VL-7B \& LLaVA-OV1.5-8B & 56.6 & 52.0 & 37.0 & 48.1\\
    \bottomrule
  \end{tabular}}
  \caption{Model agnostic. Performance of different local agents paired with a fixed coordinator agent (Qwen3-VL-8B).}
  \label{tab:model_agnostic}
\end{table*}

%% file: sec/conclusion.tex
\section{Conclusion}
We present \ours, an end-to-end multi-agent collaboration framework that enables scalable video understanding under constrained perception and communication budgets. By decomposing long videos across locally budgeted agents and introducing an agent-native latent communication protocol, \ours~decouples per-agent perception limits from global video complexity while preserving fine-grained visual semantics. A curriculum training strategy further stabilizes optimization and promotes effective cross-agent coordination in the shared communication space. 
Extensive experiments demonstrate that \ours~consistently outperforms state-of-the-art MLLMs and multi-agent baselines under identical budget constraints. Together, these results highlight compact latent multi-agent collaboration as a promising paradigm for scalable and efficient video understanding.

%% file: sec/impact.tex
\section*{Impact Statement}
This paper presents \ours, an end-to-end multi-agent collaboration framework for budget-constrained long-video understanding with multi-modal large language models (MLLMs). 
By decoupling per-agent perception budgets from global video complexity and enabling agent-native latent communication, our method can improve the scalability and robustness of video understanding under fixed compute and context constraints. 
This may benefit downstream applications that rely on long-form video perception (e.g., video QA, content understanding, and assistive analysis), by reducing information loss introduced by aggressive sampling or text-only intermediates.
Overall, this work aims to advance machine learning methods for scalable multimodal reasoning, and we do not foresee major ethical or social concerns beyond those commonly associated with improved video understanding and analysis.

%% file: example_paper.bib
@article{yin2024survey,
  title={A survey on multimodal large language models},
  author={Yin, Shukang and Fu, Chaoyou and Zhao, Sirui and Li, Ke and Sun, Xing and Xu, Tong and Chen, Enhong},
  journal={National Science Review},
  volume={11},
  number={12},
  pages={nwae403},
  year={2024},
  publisher={Oxford University Press}
}

@article{caffagni2024revolution,
  title={The revolution of multimodal large language models: a survey},
  author={Caffagni, Davide and Cocchi, Federico and Barsellotti, Luca and Moratelli, Nicholas and Sarto, Sara and Baraldi, Lorenzo and Cornia, Marcella and Cucchiara, Rita},
  journal={arXiv preprint arXiv:2402.12451},
  year={2024}
}

@article{zhang2024mm,
  title={Mm-llms: Recent advances in multimodal large language models},
  author={Zhang, Duzhen and Yu, Yahan and Dong, Jiahua and Li, Chenxing and Su, Dan and Chu, Chenhui and Yu, Dong},
  journal={arXiv preprint arXiv:2401.13601},
  year={2024}
}

@inproceedings{wang2024videoagent,
  title={Videoagent: Long-form video understanding with large language model as agent},
  author={Wang, Xiaohan and Zhang, Yuhui and Zohar, Orr and Yeung-Levy, Serena},
  booktitle={European Conference on Computer Vision},
  pages={58--76},
  year={2024},
  organization={Springer}
}

@article{videorag,
  title={Video-rag: Visually-aligned retrieval-augmented long video comprehension},
  author={Luo, Yongdong and Zheng, Xiawu and Li, Guilin and Yin, Shukang and Lin, Haojia and Fu, Chaoyou and Huang, Jinfa and Ji, Jiayi and Chao, Fei and Luo, Jiebo and others},
  journal={arXiv preprint arXiv:2411.13093},
  year={2024}
}

@inproceedings{han2024self,
  title={Self-adaptive sampling for accurate video question answering on image text models},
  author={Han, Wei and Chen, Hui and Kan, Min-Yen and Poria, Soujanya},
  booktitle={Findings of the Association for Computational Linguistics: NAACL 2024},
  pages={2522--2534},
  year={2024}
}

@article{liang2024end,
  title={End-to-end video question answering with frame scoring mechanisms and adaptive sampling},
  author={Liang, Jianxin and Meng, Xiaojun and Wang, Yueqian and Liu, Chang and Liu, Qun and Zhao, Dongyan},
  journal={arXiv preprint arXiv:2407.15047},
  year={2024}
}

@article{yao2025survey,
  title={A survey on agentic multimodal large language models},
  author={Yao, Huanjin and Zhang, Ruifei and Huang, Jiaxing and Zhang, Jingyi and Wang, Yibo and Fang, Bo and Zhu, Ruolin and Jing, Yongcheng and Liu, Shunyu and Li, Guanbin and others},
  journal={arXiv preprint arXiv:2510.10991},
  year={2025}
}

@article{yu2024frame,
  title={Frame-voyager: Learning to query frames for video large language models},
  author={Yu, Sicheng and Jin, Chengkai and Wang, Huanyu and Chen, Zhenghao and Jin, Sheng and Zuo, Zhongrong and Xu, Xiaolei and Sun, Zhenbang and Zhang, Bingni and Wu, Jiawei and others},
  journal={arXiv preprint arXiv:2410.03226},
  year={2024}
}

@article{zhou2025reagent,
  title={ReAgent-V: A Reward-Driven Multi-Agent Framework for Video Understanding},
  author={Zhou, Yiyang and He, Yangfan and Su, Yaofeng and Han, Siwei and Jang, Joel and Bertasius, Gedas and Bansal, Mohit and Yao, Huaxiu},
  journal={arXiv preprint arXiv:2506.01300},
  year={2025}
}

@article{long2025seeing,
  title={Seeing, listening, remembering, and reasoning: A multimodal agent with long-term memory},
  author={Long, Lin and He, Yichen and Ye, Wentao and Pan, Yiyuan and Lin, Yuan and Li, Hang and Zhao, Junbo and Li, Wei},
  journal={arXiv preprint arXiv:2508.09736},
  year={2025}
}

@article{zuo2025videolucy,
  title={VideoLucy: Deep Memory Backtracking for Long Video Understanding},
  author={Zuo, Jialong and Deng, Yongtai and Kong, Lingdong and Yang, Jingkang and Jin, Rui and Zhang, Yiwei and Sang, Nong and Pan, Liang and Liu, Ziwei and Gao, Changxin},
  journal={arXiv preprint arXiv:2510.12422},
  year={2025}
}

@article{kugo2025videomultiagents,
  title={VideoMultiAgents: A Multi-Agent Framework for Video Question Answering},
  author={Kugo, Noriyuki and Li, Xiang and Li, Zixin and Gupta, Ashish and Khatua, Arpandeep and Jain, Nidhish and Patel, Chaitanya and Kyuragi, Yuta and Ishii, Yasunori and Tanabiki, Masamoto and others},
  journal={arXiv preprint arXiv:2504.20091},
  year={2025}
}

@article{liu2025longvideoagent,
  title={LongVideoAgent: Multi-Agent Reasoning with Long Videos},
  author={Liu, Runtao and Liu, Ziyi and Tang, Jiaqi and Ma, Yue and Pi, Renjie and Zhang, Jipeng and Chen, Qifeng},
  journal={arXiv preprint arXiv:2512.20618},
  year={2025}
}

@article{sukhbaatar2016learning,
  title={Learning multiagent communication with backpropagation},
  author={Sukhbaatar, Sainbayar and Fergus, Rob and others},
  journal={Advances in neural information processing systems},
  volume={29},
  year={2016}
}

@article{foerster2016learning,
  title={Learning to communicate with deep multi-agent reinforcement learning},
  author={Foerster, Jakob and Assael, Ioannis Alexandros and De Freitas, Nando and Whiteson, Shimon},
  journal={Advances in neural information processing systems},
  volume={29},
  year={2016}
}

@article{alayrac2022flamingo,
  title={Flamingo: a visual language model for few-shot learning},
  author={Alayrac, Jean-Baptiste and Donahue, Jeff and Luc, Pauline and Miech, Antoine and Barr, Iain and Hasson, Yana and Lenc, Karel and Mensch, Arthur and Millican, Katherine and Reynolds, Malcolm and others},
  journal={Advances in neural information processing systems},
  volume={35},
  pages={23716--23736},
  year={2022}
}

@inproceedings{li2023blip,
  title={Blip-2: Bootstrapping language-image pre-training with frozen image encoders and large language models},
  author={Li, Junnan and Li, Dongxu and Savarese, Silvio and Hoi, Steven},
  booktitle={International conference on machine learning},
  pages={19730--19742},
  year={2023},
  organization={PMLR}
}

@article{liu2023visual,
  title={Visual instruction tuning},
  author={Liu, Haotian and Li, Chunyuan and Wu, Qingyang and Lee, Yong Jae},
  journal={Advances in neural information processing systems},
  volume={36},
  pages={34892--34916},
  year={2023}
}

@article{achiam2023gpt,
  title={Gpt-4 technical report},
  author={Achiam, Josh and Adler, Steven and Agarwal, Sandhini and Ahmad, Lama and Akkaya, Ilge and Aleman, Florencia Leoni and Almeida, Diogo and Altenschmidt, Janko and Altman, Sam and Anadkat, Shyamal and others},
  journal={arXiv preprint arXiv:2303.08774},
  year={2023}
}

@article{li2025videochat,
  title={Videochat: Chat-centric video understanding},
  author={Li, KunChang and He, Yinan and Wang, Yi and Li, Yizhuo and Wang, Wenhai and Luo, Ping and Wang, Yali and Wang, Limin and Qiao, Yu},
  journal={Science China Information Sciences},
  volume={68},
  number={10},
  pages={200102},
  year={2025},
  publisher={Springer}
}

@article{zhang2023video,
  title={Video-llama: An instruction-tuned audio-visual language model for video understanding},
  author={Zhang, Hang and Li, Xin and Bing, Lidong},
  journal={arXiv preprint arXiv:2306.02858},
  year={2023}
}

@article{chen2023videollm,
  title={Videollm: Modeling video sequence with large language models},
  author={Chen, Guo and Zheng, Yin-Dong and Wang, Jiahao and Xu, Jilan and Huang, Yifei and Pan, Junting and Wang, Yi and Wang, Yali and Qiao, Yu and Lu, Tong and others},
  journal={arXiv preprint arXiv:2305.13292},
  year={2023}
}

@article{wu2024longvideobench,
  title={Longvideobench: A benchmark for long-context interleaved video-language understanding},
  author={Wu, Haoning and Li, Dongxu and Chen, Bei and Li, Junnan},
  journal={Advances in Neural Information Processing Systems},
  volume={37},
  pages={28828--28857},
  year={2024}
}

@inproceedings{wang2025lvbench,
  title={Lvbench: An extreme long video understanding benchmark},
  author={Wang, Weihan and He, Zehai and Hong, Wenyi and Cheng, Yean and Zhang, Xiaohan and Qi, Ji and Ding, Ming and Gu, Xiaotao and Huang, Shiyu and Xu, Bin and others},
  booktitle={Proceedings of the IEEE/CVF International Conference on Computer Vision},
  pages={22958--22967},
  year={2025}
}

@inproceedings{wu2024autogen,
  title={Autogen: Enabling next-gen LLM applications via multi-agent conversations},
  author={Wu, Qingyun and Bansal, Gagan and Zhang, Jieyu and Wu, Yiran and Li, Beibin and Zhu, Erkang and Jiang, Li and Zhang, Xiaoyun and Zhang, Shaokun and Liu, Jiale and others},
  booktitle={First Conference on Language Modeling},
  year={2024}
}

@article{li2023camel,
  title={Camel: Communicative agents for" mind" exploration of large language model society},
  author={Li, Guohao and Hammoud, Hasan and Itani, Hani and Khizbullin, Dmitrii and Ghanem, Bernard},
  journal={Advances in Neural Information Processing Systems},
  volume={36},
  pages={51991--52008},
  year={2023}
}

@inproceedings{qian2024chatdev,
  title={Chatdev: Communicative agents for software development},
  author={Qian, Chen and Liu, Wei and Liu, Hongzhang and Chen, Nuo and Dang, Yufan and Li, Jiahao and Yang, Cheng and Chen, Weize and Su, Yusheng and Cong, Xin and others},
  booktitle={Proceedings of the 62nd Annual Meeting of the Association for Computational Linguistics (Volume 1: Long Papers)},
  pages={15174--15186},
  year={2024}
}

@inproceedings{hong2023metagpt,
  title={MetaGPT: Meta programming for a multi-agent collaborative framework},
  author={Hong, Sirui and Zhuge, Mingchen and Chen, Jonathan and Zheng, Xiawu and Cheng, Yuheng and Wang, Jinlin and Zhang, Ceyao and Wang, Zili and Yau, Steven Ka Shing and Lin, Zijuan and others},
  booktitle={The Twelfth International Conference on Learning Representations},
  year={2023}
}

@inproceedings{fan2024videoagent,
  title={Videoagent: A memory-augmented multimodal agent for video understanding},
  author={Fan, Yue and Ma, Xiaojian and Wu, Rujie and Du, Yuntao and Li, Jiaqi and Gao, Zhi and Li, Qing},
  booktitle={European Conference on Computer Vision},
  pages={75--92},
  year={2024},
  organization={Springer}
}

@article{wu2023visual,
  title={Visual chatgpt: Talking, drawing and editing with visual foundation models},
  author={Wu, Chenfei and Yin, Shengming and Qi, Weizhen and Wang, Xiaodong and Tang, Zecheng and Duan, Nan},
  journal={arXiv preprint arXiv:2303.04671},
  year={2023}
}

@article{shen2023hugginggpt,
  title={Hugginggpt: Solving ai tasks with chatgpt and its friends in hugging face},
  author={Shen, Yongliang and Song, Kaitao and Tan, Xu and Li, Dongsheng and Lu, Weiming and Zhuang, Yueting},
  journal={Advances in Neural Information Processing Systems},
  volume={36},
  pages={38154--38180},
  year={2023}
}

@article{yang2023mm,
  title={Mm-react: Prompting chatgpt for multimodal reasoning and action},
  author={Yang, Zhengyuan and Li, Linjie and Wang, Jianfeng and Lin, Kevin and Azarnasab, Ehsan and Ahmed, Faisal and Liu, Zicheng and Liu, Ce and Zeng, Michael and Wang, Lijuan},
  journal={arXiv preprint arXiv:2303.11381},
  year={2023}
}

@article{dvd,
  title={Deep Video Discovery: Agentic Search with Tool Use for Long-form Video Understanding},
  author={Zhang, Xiaoyi and Jia, Zhaoyang and Guo, Zongyu and Li, Jiahao and Li, Bin and Li, Houqiang and Lu, Yan},
  journal={arXiv preprint arXiv:2505.18079},
  year={2025}
}

@article{su2025thinking,
  title={Thinking with images for multimodal reasoning: Foundations, methods, and future frontiers},
  author={Su, Zhaochen and Xia, Peng and Guo, Hangyu and Liu, Zhenhua and Ma, Yan and Qu, Xiaoye and Liu, Jiaqi and Li, Yanshu and Zeng, Kaide and Yang, Zhengyuan and others},
  journal={arXiv preprint arXiv:2506.23918},
  year={2025}
}

@article{fu2025cache,
  title={Cache-to-Cache: Direct Semantic Communication Between Large Language Models},
  author={Fu, Tianyu and Min, Zihan and Zhang, Hanling and Yan, Jichao and Dai, Guohao and Ouyang, Wanli and Wang, Yu},
  journal={arXiv preprint arXiv:2510.03215},
  year={2025}
}

@article{zou2025latent,
  title={Latent collaboration in multi-agent systems},
  author={Zou, Jiaru and Yang, Xiyuan and Qiu, Ruizhong and Li, Gaotang and Tieu, Katherine and Lu, Pan and Shen, Ke and Tong, Hanghang and Choi, Yejin and He, Jingrui and others},
  journal={arXiv preprint arXiv:2511.20639},
  year={2025}
}

@article{ge2023context,
  title={In-context autoencoder for context compression in a large language model},
  author={Ge, Tao and Hu, Jing and Wang, Lei and Wang, Xun and Chen, Si-Qing and Wei, Furu},
  journal={arXiv preprint arXiv:2307.06945},
  year={2023}
}

@article{he2025clara,
  title={CLaRa: Bridging Retrieval and Generation with Continuous Latent Reasoning},
  author={He, Jie and Bai, Richard He and Williamson, Sinead and Pan, Jeff Z and Jaitly, Navdeep and Zhang, Yizhe},
  journal={arXiv preprint arXiv:2511.18659},
  year={2025}
}

@article{mozaffari2024slim,
  title={SLiM: One-shot Quantization and Sparsity with Low-rank Approximation for LLM Weight Compression},
  author={Mozaffari, Mohammad and Yazdanbakhsh, Amir and Dehnavi, Maryam Mehri},
  journal={arXiv preprint arXiv:2410.09615},
  year={2024}
}

@article{frantar2022gptq,
  title={Gptq: Accurate post-training quantization for generative pre-trained transformers},
  author={Frantar, Elias and Ashkboos, Saleh and Hoefler, Torsten and Alistarh, Dan},
  journal={arXiv preprint arXiv:2210.17323},
  year={2022}
}

@misc{qwen3,
      title={Qwen3 Technical Report}, 
      author={Qwen Team},
      year={2025},
      eprint={2505.09388},
      archivePrefix={arXiv},
      primaryClass={cs.CL},
      url={https://arxiv.org/abs/2505.09388}, 
}

@article{qwen2.5,
  title={Qwen2. 5-vl technical report},
  author={Bai, Shuai and Chen, Keqin and Liu, Xuejing and Wang, Jialin and Ge, Wenbin and Song, Sibo and Dang, Kai and Wang, Peng and Wang, Shijie and Tang, Jun and others},
  journal={arXiv preprint arXiv:2502.13923},
  year={2025}
}

@article{gpt-4o,
  title={Gpt-4o system card},
  author={Hurst, Aaron and Lerer, Adam and Goucher, Adam P and Perelman, Adam and Ramesh, Aditya and Clark, Aidan and Ostrow, AJ and Welihinda, Akila and Hayes, Alan and Radford, Alec and others},
  journal={arXiv preprint arXiv:2410.21276},
  year={2024}
}

@article{video-r1,
  title={Video-r1: Reinforcing video reasoning in mllms},
  author={Feng, Kaituo and Gong, Kaixiong and Li, Bohao and Guo, Zonghao and Wang, Yibing and Peng, Tianshuo and Wu, Junfei and Zhang, Xiaoying and Wang, Benyou and Yue, Xiangyu},
  journal={arXiv preprint arXiv:2503.21776},
  year={2025}
}

@misc{zhang2024llavanextvideo,
  title={LLaVA-NeXT: A Strong Zero-shot Video Understanding Model},
  url={https://llava-vl.github.io/blog/2024-04-30-llava-next-video/},
  author={Zhang, Yuanhan and Li, Bo and Liu, haotian and Lee, Yong jae and Gui, Liangke and Fu, Di and Feng, Jiashi and Liu, Ziwei and Li, Chunyuan},
  month={April},
  year={2024}
}

@misc{llava-video-178k,
    title={Video Instruction Tuning With Synthetic Data}, 
    author={Yuanhan Zhang and Jinming Wu and Wei Li and Bo Li and Zejun Ma and Ziwei Liu and Chunyuan Li},
    year={2024},
    eprint={2410.02713},
    archivePrefix={arXiv},
    primaryClass={cs.CV},
    url={https://arxiv.org/abs/2410.02713}, 
}

@article{chen2024sharegpt4video,
  title={Sharegpt4video: Improving video understanding and generation with better captions},
  author={Chen, Lin and Wei, Xilin and Li, Jinsong and Dong, Xiaoyi and Zhang, Pan and Zang, Yuhang and Chen, Zehui and Duan, Haodong and Tang, Zhenyu and Yuan, Li and others},
  journal={Advances in Neural Information Processing Systems},
  volume={37},
  pages={19472--19495},
  year={2024}
}

@article{clark2026molmo2,
  title={Molmo2: Open Weights and Data for Vision-Language Models with Video Understanding and Grounding},
  author={Clark, Christopher and Zhang, Jieyu and Ma, Zixian and Park, Jae Sung and Salehi, Mohammadreza and Tripathi, Rohun and Lee, Sangho and Ren, Zhongzheng and Kim, Chris Dongjoo and Yang, Yinuo and others},
  journal={arXiv preprint arXiv:2601.10611},
  year={2026}
}

@article{liu2024kangaroo,
  title={Kangaroo: A powerful video-language model supporting long-context video input},
  author={Liu, Jiajun and Wang, Yibing and Ma, Hanghang and Wu, Xiaoping and Ma, Xiaoqi and Wei, Xiaoming and Jiao, Jianbin and Wu, Enhua and Hu, Jie},
  journal={arXiv preprint arXiv:2408.15542},
  year={2024}
}

@article{videollama3,
  title={Videollama 3: Frontier multimodal foundation models for image and video understanding},
  author={Zhang, Boqiang and Li, Kehan and Cheng, Zesen and Hu, Zhiqiang and Yuan, Yuqian and Chen, Guanzheng and Leng, Sicong and Jiang, Yuming and Zhang, Hang and Li, Xin and others},
  journal={arXiv preprint arXiv:2501.13106},
  year={2025}
}

@article{videollama2,
  title={Videollama 2: Advancing spatial-temporal modeling and audio understanding in video-llms},
  author={Cheng, Zesen and Leng, Sicong and Zhang, Hang and Xin, Yifei and Li, Xin and Chen, Guanzheng and Zhu, Yongxin and Zhang, Wenqi and Luo, Ziyang and Zhao, Deli and others},
  journal={arXiv preprint arXiv:2406.07476},
  year={2024}
}

@article{an2025llava,
  title={Llava-onevision-1.5: Fully open framework for democratized multimodal training},
  author={An, Xiang and Xie, Yin and Yang, Kaicheng and Zhang, Wenkang and Zhao, Xiuwei and Cheng, Zheng and Wang, Yirui and Xu, Songcen and Chen, Changrui and Zhu, Didi and others},
  journal={arXiv preprint arXiv:2509.23661},
  year={2025}
}

@article{keye1.5,
  title={Kwai keye-vl 1.5 technical report},
  author={Yang, Biao and Wen, Bin and Ding, Boyang and Liu, Changyi and Chu, Chenglong and Song, Chengru and Rao, Chongling and Yi, Chuan and Li, Da and Zang, Dunju and others},
  journal={arXiv preprint arXiv:2509.01563},
  year={2025}
}

@article{internvl2.5,
  title={Expanding performance boundaries of open-source multimodal models with model, data, and test-time scaling},
  author={Chen, Zhe and Wang, Weiyun and Cao, Yue and Liu, Yangzhou and Gao, Zhangwei and Cui, Erfei and Zhu, Jinguo and Ye, Shenglong and Tian, Hao and Liu, Zhaoyang and others},
  journal={arXiv preprint arXiv:2412.05271},
  year={2024}
}

@article{MapReduce,
  title={Mr. video:" mapreduce" is the principle for long video understanding},
  author={Pang, Ziqi and Wang, Yu-Xiong},
  journal={arXiv preprint arXiv:2504.16082},
  year={2025}
}

@article{LatentMAS,
  title={Latent collaboration in multi-agent systems},
  author={Zou, Jiaru and Yang, Xiyuan and Qiu, Ruizhong and Li, Gaotang and Tieu, Katherine and Lu, Pan and Shen, Ke and Tong, Hanghang and Choi, Yejin and He, Jingrui and others},
  journal={arXiv preprint arXiv:2511.20639},
  year={2025}
}

@inproceedings{video-mme,
  title={Video-mme: The first-ever comprehensive evaluation benchmark of multi-modal llms in video analysis},
  author={Fu, Chaoyou and Dai, Yuhan and Luo, Yongdong and Li, Lei and Ren, Shuhuai and Zhang, Renrui and Wang, Zihan and Zhou, Chenyu and Shen, Yunhang and Zhang, Mengdan and others},
  booktitle={Proceedings of the Computer Vision and Pattern Recognition Conference},
  pages={24108--24118},
  year={2025}
}

@inproceedings{zhou2025mlvu,
  title={Mlvu: Benchmarking multi-task long video understanding},
  author={Zhou, Junjie and Shu, Yan and Zhao, Bo and Wu, Boya and Liang, Zhengyang and Xiao, Shitao and Qin, Minghao and Yang, Xi and Xiong, Yongping and Zhang, Bo and others},
  booktitle={Proceedings of the Computer Vision and Pattern Recognition Conference},
  pages={13691--13701},
  year={2025}
}
